\documentclass{article}

\usepackage[preprint]{neurips_2026}

\usepackage[utf8]{inputenc} 
\usepackage[T1]{fontenc}    
\usepackage{hyperref}       
\usepackage{url}            
\usepackage{booktabs}       
\usepackage{amsmath}        
\usepackage{amsfonts}       
\usepackage{graphicx}       
\usepackage{nicefrac}       
\usepackage{microtype}      
\usepackage{xcolor}         
\usepackage{wrapfig}
\usepackage{color, colortbl}
\usepackage{array} 
\usepackage{adjustbox}
\usepackage{multirow}
\usepackage{xspace}

\usepackage{todonotes}
\newcommand{\footnotescriptsize}{\fontsize{8pt}{9pt}\selectfont}

\makeatletter
\DeclareRobustCommand\onedot{\futurelet\@let@token\@onedot}
\def\@onedot{\ifx\@let@token.\else.\null\fi\xspace}

\def\eg{\emph{e.g}\onedot} 
\def\ie{\emph{i.e}\onedot}

\makeatother

\definecolor{gg}{gray}{0.70}

\definecolor{baselinecolor}{gray}{.9}
\newcommand{\cc}[1]{\cellcolor{baselinecolor}{#1}}

\title{PAI: Preserving Amplitude Information in Representation-Based Time-Series Anomaly Detection}

\author{
Kang Zhang$^1~^2$ \quad
Wei Jian Lau$^1$ \quad
Shoushou Ren$^1$ \quad 
Dong Lin$^1$ \quad
\\
\textbf{
Joon Son Chung$^2$ \quad
Chuanhao Sun$^1$}\thanks{Corresponding author.}  \\
$^1$HUAWEI \quad
$^2$KAIST \\
{\tt\small zhangkang@kaist.ac.kr, 	sun.chuanhao@huawei.com} \\
}

\begin{document}

\maketitle

\begin{abstract}
Representation-based time-series anomaly detection algorithms significantly outperform other methods on diverse anomaly detection tasks. However, we notice that they suffer from a major limitation in our evaluation --- \textbf{their learned embeddings are often amplitude-agnostic}. Lossing amplitude information can degrade performance on amplitude related anomalies, and this failure is prevalent across all existing representation-based methods. To address aforementioned issues, we propose a new anomaly scoring scheme named PAI. PAI consists of two complementary modules, a diagnostic module and a final score augmentation function. The diagnostic module compares cosine and Euclidean scoring on the same representation bank to test whether amplitude information is already captured in the learned representation. Then in final score augmentation function, PAI computes a point-wise median/MAD deviation score and a local mean-shift score—which are fused with the representation score to produce the final anomaly score. 
On the TSB-AD-U-Eva and TAB UV datasets, PAI improves all four evaluated representation-based methods across every reported metric, achieving average VUS-PR gains of 98.4\% and 36.8\%, respectively. Among all evaluated combinations, PaAno + PAI achieves the best performance, outperforming the state-of-the-art method by 15\%.
Further evaluation on bootstrap confidence intervals, anomaly-type breakdowns, and a TS2Vec input-normalization ablation further support the proposed scheme. These results suggest that explicitly retaining amplitude information is important for representation-based time-series anomaly detection, which has been underemphasized in existing scoring schemes. 
{Code is available at: \url{https://github.com/pantheon5100/PAI}}
\end{abstract}

\section{Introduction}
\label{sec:introduction}

Time-series anomaly detection (TSAD)  is a critical machine learning (ML) task with broad applications such as financial fraud detection, network diagnostics, rare disease detection, and social media analysis. Recent works have shown that representation-based methods~\cite{park2026paano, yue2022ts2vec, yang2023dcdetector, ekambaram2026tspulse} outperform many existing approaches across a wide range of TSAD domains. These methods learn representative embeddings using contrastive encoders, patch-memory models, or pretrained time-series backbones, enabling them to capture temporal context around a patch beyond raw pointwise deviations. As a result, they are particular effective at identifying contextual and pattern anomalies, which are often difficult to detect using simple pointwise rules.

Although representation-based TASD methods perform well in many settings, we identify a severe failure mode: \textbf{amplitude information may be suppressed or ignored in the final anomaly score}. The failure can arise at different stages of the algorithm. At the encoder stage, input normalization may remove level or amplitude information before the representation is learned. At the scoring stage, cosine distance, normalized attention, or native-head may expose only part of the learned state, preventing amplitude information from contributing effectively to the final score. Consequently, the same amplitude-related anomaly can be ranked very differently depending on whether the detector preserves amplitude information in its scoring function. This limitation is significant because time-series anomalies can manifest as spikes, level shifts, intensity changes, variance changes, or other amplitude-dependent variations. Effective TSAD therefore requires the detector to preserve both raw-signal scale and learned temporal structure.

In this work, we introduce PAI, a lightweight fusion scoring scheme designed to preserve amplitude information in representation-based TSAD. We first motivate amplitude-aware scoring through a diagnostic study on TSB-AD-U-Eva dataset, where we compare cosine- and Euclidean-based anomaly scores computed over the same normal representation bank. This test evaluates whether amplitude-related information is present in the learned representation. The gains from Euclidean scoring vary across detection methods, indicating that some representations preserve useful amplitude information while others remain largely amplitude-agnostic. Based on this observation, PAI replaces the original cosine-distance representation score with a Euclidean-distance score, and further augments it with two train-calibrated amplitude scores: a pointwise median/MAD deviation score, \texttt{magG}, and a local mean-shift score, \texttt{T2}. PAI is designed as a score-level extension to existing representation-based methods and requires no changes to their architectures or training procedures. 

\begin{wrapfigure}{r}{0.48\linewidth}
    \vspace{-12pt}
    \centering
    \includegraphics[width=1.0\linewidth]{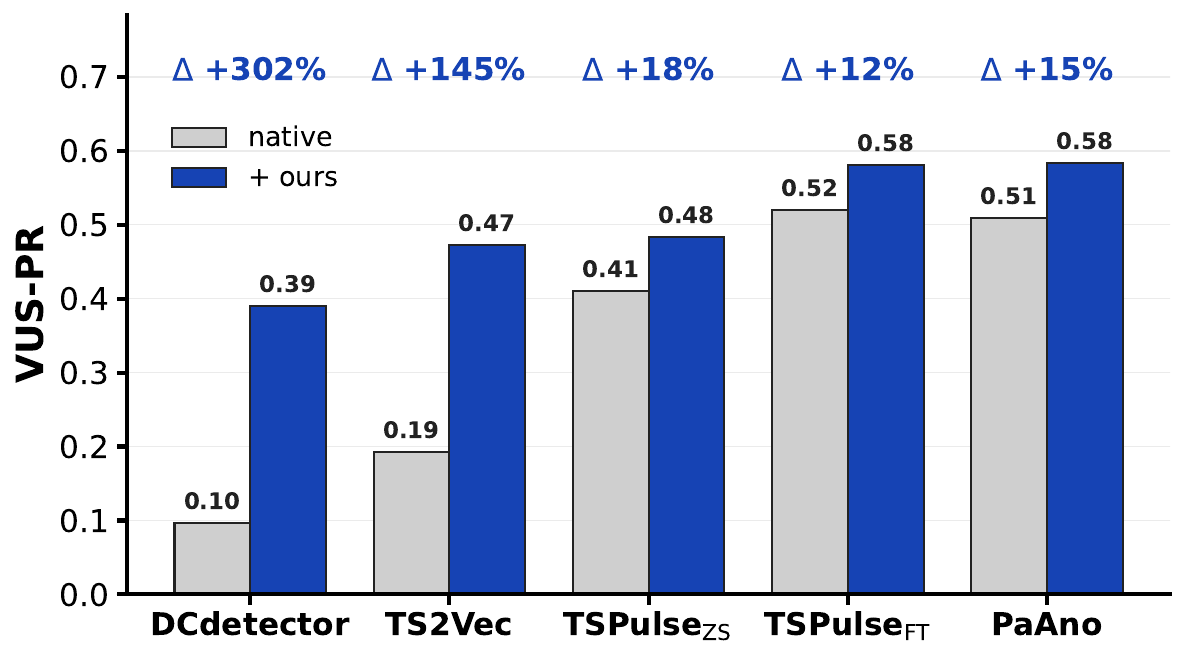}
    \vspace{-13pt}
    \caption{
    \textbf{\texttt{PAI} uniformly improves representation-based TSAD methods on TSB-AD-U-Eva~\cite{liu2024the}.}
    }
    \label{fig:teaser}
    \vspace{-15pt}
\end{wrapfigure}

On TSB-AD-U-Eva~\cite{liu2024the}, augmenting the scoring pipeline of existing representation-based detectors with PAI yields significant anomaly detection improvements. As shown in Figure~\ref{fig:teaser}, PAI improves DCdetector, TS2Vec, TSPulse zero-shot, and PaAno~\cite{park2026paano} across all reported metrics. In particular, \textbf{PaAno + PAI} achieves the highest VUS-PR on the benchmark, while fine-tuned TSPulse equipped with the same amplitude scores obtains a nearly tied result. We further evaluate the robustness of these gains through fixed-score transfer to TAB UV, paired-bootstrap comparisons against classical baselines, anomaly-type breakdowns, a TS2Vec input-RevIN ablation, weight-robustness checks, and compute-cost measurements.

Our contributions are:
\begin{itemize}
    \item We identify a critical limitation of existing univariate representation-based TSAD methods: they fail to preserve amplitude information, resulting in nearly amplitude-invariant detection behavior.

    \item We introduce PAI, a lightweight, scoring scheme augmentation that combines a Euclidean representation score with pointwise and window-scale raw-amplitude scores.

    \item A cosine-versus-Euclidean diagnostic to test whether learned representations already encode amplitude information, providing a simple indicator of when explicit raw-amplitude correction is needed.

    \item A thorough evaluation on representative TSAD benchmarks is conducted, showing that PAI consistently improves existing representation-based detectors and achieves a 15\% relative VUS-PR gain, improving performance from \textbf{0.51 to 0.58} on the state-of-the-art method.
\end{itemize}

\section{Related Work}
\paragraph{Raw-signal and residual-based methods.}
Classical statistical methods~\cite{charu2017pca,papa2017kshape2, zeng2023dlinear}, distance-based methods~\cite{iglewicz1993volume, peter1993alternatives, Aminikhanghahi2017}, and forecasting or
reconstruction models~\cite{ekambaram2026tspulse} often score anomalies through deviations in the observed
signal or its prediction residual. This makes them naturally sensitive to
spikes, level shifts, variance changes, and other raw-amplitude events. Neural
models such as Anomaly Transformer~\citep{xu2022anomaly} and
TranAD~\citep{tuli2022tranad} extend this
idea with higher-capacity temporal modeling and derive anomaly scores from
reconstruction, prediction, or attention-discrepancy behavior. These methods are
important because they show that raw-signal deviations can be central in TSAD.
However, their representation learning and scoring rules are usually
coupled inside the same detector, making it difficult to isolate whether a gain
comes from better temporal representation, better raw-amplitude sensitivity, or
the final score formulation.

\paragraph{Contrastive and metric representation-based TSAD.}
Another line of TSAD methods formulates anomaly detection as representation learning~\cite{chen2020simple,He_2020_CVPR,zhang2022dual,caron2021emerging} followed by scoring in a learned embedding space. TS2Vec~\cite{yue2022ts2vec} learns timestamp-level and subseries-level representations through hierarchical contrastive objectives, and applies representation dissimilarity for downstream anomaly detection. DCdetector~\citep{yang2023dcdetector} learns contrastive representations with a dual-attention mechanism and detects anomalies based on discrepancies between learned representation views. PaAno~\citep{park2026paano} encodes time-series patches and scores each test patch by comparing its representation with a memory bank of normal patterns. Our work targets this class of representation-based methods. We show that, despite their strong detection performance, their learned embeddings are often amplitude-agnostic, causing amplitude-related anomalies to be under-scored after normalization and encoding. PAI addresses this shared limitation as a plug-in scoring scheme: it diagnoses amplitude preservation by comparing cosine and Euclidean scores on the same representation bank, and then augments the representation score with raw-signal median/MAD and local mean-shift deviations.

\paragraph{Pretrained and multi-head time-series models.}
Recent pretrained and multi-head time-series models expand the representation choices for TSAD. Foundation models such as MOMENT~\citep{goswami2024moment} and TimesFM~\citep{das2024timesfm} learn transferable temporal representations across tasks, while TSPulse learns compact disentangled representations for efficient time-series analysis~\citep{ekambaram2026tspulse}. We discuss these models separately from patch-memory and metric-based detectors because their anomaly scores are not necessarily based only on embedding comparisons. For multi-head models, amplitude information may enter the final score through prediction, reconstruction, or other task-specific errors, even if some internal representations are normalized. Thus, this family provides a broader test case for our diagnosis: PAI evaluates whether the final scoring pipeline preserves amplitude information, regardless of whether the detector uses memory-bank matching, pretrained representations, or native scoring heads.

\begin{figure}[t]
    \centering
    \includegraphics[width=\textwidth]{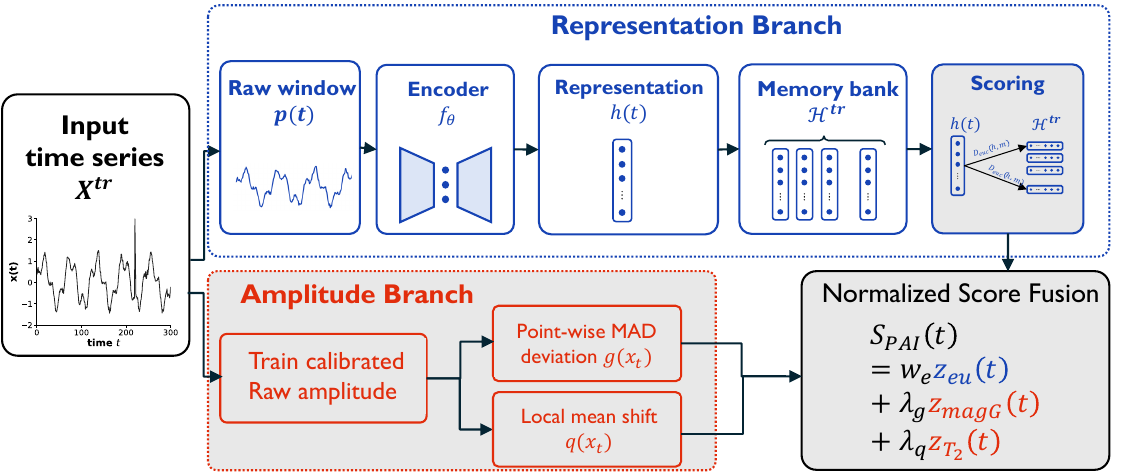}
    \caption{\textbf{PAI workflow.} PAI augments the scoring scheme of exsiting representation-based TSAD methods with two parallel branches. PAI replaces the cosine distance with Euclidean distance in the representation scoring function branch. Concurrently, it computes two more train calibrated raw amplitude scores which are then fused with the score from the representation branch to form the final anomaly score. Modules with colored background show new or modified components in the scoring scheme introduced by PAI, while the other modules are unchanged from the original TSAD method.}
    \label{fig:method}
    \vspace{-10pt}
\end{figure}

\section{Method}
\label{sec:method}

We propose PAI, an anomaly scoring augmentation scheme for representation-based anomaly detection in univariate time series. PAI preserves the base representation-learning model and refines only its final anomaly scores. It uses contextual and shape cues encoded by the learned representation, while explicitly supplementing them with raw-amplitude evidence that may be attenuated during representation learning. Specifically, PAI augments the base anomaly score by replacing cosine distance scores with Euclidean distance scores, then two additional raw-signal scores calibrated only from normal training data. Figure~\ref{fig:method} illustrates the proposed method. In this section, we first formulate the representation-based scoring setting, then introduce a cosine-Euclidean diagnostic for analyzing the significance of radial information in the embedding space, and finally define the fixed score-fusion rule used for anomaly detection.

\subsection{Setting and base representation score}
\label{sec:problem_setting}
\paragraph{Training data and patches.}
Let $X^{\mathrm{tr}}=(x^{\mathrm{tr}}_1,\ldots,x^{\mathrm{tr}}_n)$ be a normal training sequence and $X=(x_1,\ldots,x_T)$ be a test sequence. The goal is to produce a score $S(t)$ for each timestamp $t$, where higher scores indicate more anomalous observations. 

A representation-based method maps a local temporal patch $p=(x_i,...,x_{i+w-1})$, \ie overlapping subsequences extracted from $X$ with a window size $w$, to an embedding $h=f_\theta(p)\in\mathbb{R}^d$ with a trained representation encoder $f_\theta$. This representation embeddings server as anomaly detection unit for representation based method for TSAD.

\paragraph{Memory bank.}
Let $\mathcal{P}_t=\{p_t\}^{N-w+1}_{t=1}$ denote the set of all the patches for a giving time series $X$. For bank-compatible encoders, we build a normal memory bank from training patches $\mathcal{P}_t^{tr}$ extracted from data $X^{\mathrm{tr}}$, $\mathcal{H}^{\mathrm{tr}}=\{f_\theta(p):p\in\mathcal{P}^{\mathrm{tr}}\}$.
Following the PaAno memory protocol~\cite{park2026paano}, we cluster $\mathcal{H}^{\mathrm{tr}}$ into $K$ clusters and keep the $K$ embeddings that closest to each cluster centroid, resulting a reduced memory bank $\hat{\mathcal{H}}^{\mathrm{tr}}=\{m_1,\ldots,m_K\}\subset\mathcal{H}^{\mathrm{tr}}$. This memory bank serves as the oracle samples for anomaly detection.

\paragraph{Representation anomaly scoring.} For a time series $X$, we first get the patches $\mathcal{P}$ with fixed size $w$ and unit stride.
Given a distance function $D(h,m)$, \eg cosine distance $D_{cos}(h,m)=1-\frac{h^\top m}{(\|h\|_2+\epsilon)(\|m\|_2+\epsilon)}$ or Euclidean distance $D_{Euc}(h,m)=\|h-m \|$, the patch-level bank score is
\begin{equation}
    S_p(p)=\frac{1}{k}\sum_{m\in\hat{\mathcal{H}}^{\mathrm{tr}}}D(f_\theta(p),m),
    \label{eq:patch_bank_score}
\end{equation}
where $\hat{\mathcal{H}}^{\mathrm{tr}}$ is the reduced memory bank on the training data $X^{tr}$ and $p\in\mathcal{P}$.
And the timestamp-level score is
\begin{equation}
    S(t)=\frac{1}{|\mathcal{P}_t|}\sum_{p\in\mathcal{P}_t}S_p(p).
    \label{eq:timestamp_bank_score}
\end{equation}
Rather than evaluating $x_{t}$ in isolation, representation scoring branch assigns an anomaly score by aggregating information from the patches surrounding timestamp $t$. A large value of $S(t)$ indicates that this local temporal neighborhood differs markedly from the normal patterns observed in the training dataset $X^{tr}$, and therefore signals a potential anomaly.

\begin{wrapfigure}{r}{0.48\linewidth}
    \vspace{-36pt}
    \centering
    \includegraphics[width=1.0\linewidth]{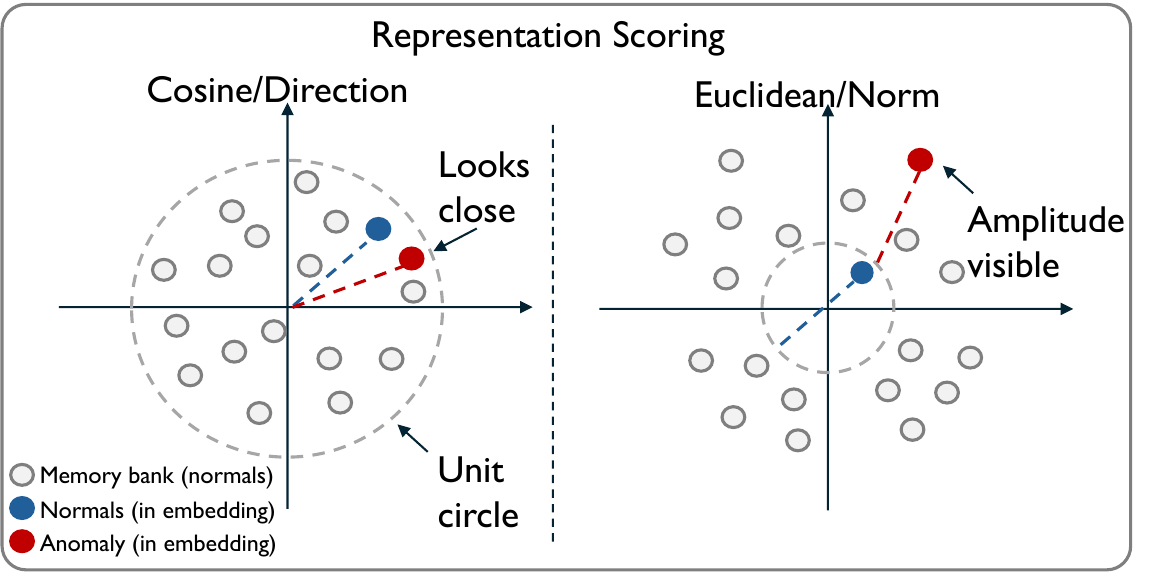}
    \vspace{-13pt}
    \caption{\textbf{Cosine versus Euclidean scoring.}
Cosine compares embedding direction, while Euclidean also captures norm, preserving amplitude information when encoded radially.}
    \label{fig:score_geometry}
    \vspace{-12pt}
\end{wrapfigure}

\subsection{Cosine-Euclidean diagnostic}
\label{sec:representation_score}

Current representation-based TSAD methods, such as PaAno~\cite{park2026paano}, typically compute the final anomaly score using the cosine distance between the current patch embedding and its nearest neighbors in a normal representation bank. Mathematically, the cosine distance between an embedding $h$ and a bank element $m$ as in Equation~\ref{eq:cos_bank}.

Cosine scoring emphasizes angular mismatch and suppresses radial mismatch. In contrast, Euclidean distance retains both angular and radial components as in Equation~\ref{eq:eu_cos_relation}.

We use a cosine–Euclidean diagnostic to test whether representation-bank scoring benefits from radial information in the embedding space. The comparison keeps the encoder $f_\theta$, training bank $\mathcal{H}^{\text{tr}}$, nearest-neighbor count, patch-to-timestamp aggregation, and training-score normalization fixed, changing only the distance function used to compute the final anomaly score. As illustrated in Figure~\ref{fig:score_geometry}, two patches may have embeddings with similar directions but different amplitude. In this case, cosine distance can remain small because it is insensitive to embedding norm, whereas Euclidean distance can separate the patches by preserving radial differences. Thus, improvements from Euclidean scoring indicate that useful amplitude-related information is encoded in the representation norm but is suppressed by cosine-based scoring.

\begin{equation}
    D_{\cos}(h,m)=1-\frac{h^\top m}{(\|h\|_2+\epsilon)(\|m\|_2+\epsilon)} .
    \label{eq:cos_bank}
\end{equation}

\begin{equation}
    D_{euc}(h,m)=\|h-m\|_2^2=\underbrace{\|h\|_2^2+\|m\|_2^2}_{\text{radial terms}}-2\underbrace{\|h\|_2\|m\|_2\cos\angle(h,m)}_{\text{angular--radial interaction}}.
    \label{eq:eu_cos_relation}
\end{equation}

Let $c(t)$ and $e(t)$ denote the representation-bank scores obtained with cosine and Euclidean distances, respectively. A large empirical gain from $e(t)$ over $c(t)$ indicates that information discarded by angular scoring is useful for anomaly ranking. Conversely, a small gain suggests that the representation score is effectively angular, or that embedding norm does not contain useful anomaly evidence. This diagnostic does not assume that embedding norm directly equals raw amplitude; it only tests whether radial information is useful for scoring. In Section~\ref{sec:diagnostic_results}, we show that replacing cosine distance with Euclidean distance in the representation-based anomaly scoring pipeline can improve the anomaly score curve even for encoders trained with cosine-based objectives.

\subsection{Raw-amplitude fusion}
\label{sec:operational_score_section}
Although representation-based scores can capture contextual and shape-level deviations, they may not fully preserve raw-amplitude information. To examine this, we compute the Spearman rank correlation~\cite{10.1093/ije/dyq191} between the base anomaly score and two raw-amplitude scores on TSB-AD-U-Eva dataset~\cite{liu2024the}. As shown in Table~\ref{tab:correlation_native_vs_mag}, most correlations are close to zero, suggesting that representation-space scores and raw-amplitude scores capture complementary aspects of anomalous behavior. The positive correlation between TS2Vec~\cite{yue2022ts2vec} and \texttt{magG} further indicates that some encoders may already encode pointwise amplitude to a degree, but this behavior is not consistent across methods. PAI therefore augments the base anomaly score with two raw-signal scores calibrated only from the normal training sequence.

\begin{table}[t]
\centering
\small
\caption{
\textbf{Spearman rank correlation $|\rho|$ between the base anomaly score and raw-amplitude scores on TSB-AD-U-Eva~\cite{liu2024the}.} For each time-series instance $i$, we compute Spearman's $\rho$ between the aligned test-time score sequences $\{b_i(t)\}_{t\in\mathcal{T}_i}$ and $\{r_i(t)\}_{t\in\mathcal{T}_i}$, where $r_i\in\{g_i,q_i,z_{g,i}+z_{q,i}\}$, and report the mean correlation across instances. Low $|\rho|$ indicates that raw-scale information is largely complementary to the base score. VUS-PR and PAI improvement are reported for context.
}
\label{tab:correlation_native_vs_mag}
\begin{adjustbox}{max width=\textwidth}
\begin{tabular}{lccccc}
\toprule
Base method & VUS-PR & + PAI $\Delta$ & $\rho(\cdot,\texttt{magG})$ & $\rho(\cdot,\mathrm{T2})$ & $\rho(\cdot,\texttt{magG}+\mathrm{T2})$ \\
\midrule
DCdetector~\citep{yang2023dcdetector} & 0.097 & $+0.293$ & $-0.086$ & $-0.032$ & $-0.084$ \\
PaAno cosine~\citep{park2026paano} & 0.509 & $+0.075$ & $+0.044$ & $+0.063$ & $+0.054$ \\
TSPulse$_{\mathrm{ZS}}$~\citep{ekambaram2026tspulse} & 0.411 & $+0.054$ & $+0.135$ & $+0.163$ & $+0.185$ \\
TS2Vec~\citep{yue2022ts2vec} & 0.193 & $+0.280$ & $\mathbf{+0.375}$ & $-0.007$ & $+0.256$ \\
\bottomrule
\end{tabular}
\end{adjustbox}
\vspace{-12pt}
\end{table}

Let
\begin{equation}
    \mu_x=\operatorname{median}(X^{\mathrm{tr}}), \qquad
    \sigma_x=\operatorname{MAD}(X^{\mathrm{tr}})+\epsilon ,
\end{equation}
where $\operatorname{MAD}$ denotes the median absolute deviation. The first raw-scale score, denoted by \texttt{magG}, measures pointwise amplitude deviation:
\begin{equation}
    g(t)=\frac{|x_t-\mu_x|}{\sigma_x}.
    \label{eq:magg}
\end{equation}
This score targets spikes and isolated threshold-crossing deviations. The second raw-scale score, denoted by T2, captures local level shifts. Given an offline radius-$W$ window
\begin{equation}
    \mathcal{W}_t=\{\tau:\max(1,t-W)\leq\tau\leq\min(T,t+W)\},
\end{equation}
we define
\begin{equation}
    q(t)=\frac{\left||\mathcal{W}_t|^{-1}\sum_{\tau\in\mathcal{W}_t}x_\tau-\mu_x\right|}{\sigma_x}.
    \label{eq:t2}
\end{equation}
We use $W=32$ unless otherwise stated. The symmetric window is used for the offline benchmark setting; for causal deployment, it can be replaced with a trailing window.

The representation based score $S_t(t)$, amplitude based pointwise score \texttt{magG} $g(t)$, and local scores $q(t)$ have different numerical scales itself. PAI therefore standardizes each score using statistics computed on the same normal training sequence. Let $S_t^{\mathrm{tr}}(u)$, $g^{\mathrm{tr}}(u)$, and $q^{\mathrm{tr}}(u)$ denote the corresponding scores evaluated at training timestamp $u$. We define
\begin{equation}
    z_{euc}(t)=\frac{S_t(t)-\mu_{S_t}^{\mathrm{tr}}}{\sigma_{S_t}^{\mathrm{tr}}+\epsilon}, 
    \qquad
    z_{magG}(t)=\frac{g(t)-\mu_g^{\mathrm{tr}}}{\sigma_g^{\mathrm{tr}}+\epsilon},
    \qquad
    z_{T_2}(t)=\frac{q(t)-\mu_q^{\mathrm{tr}}}{\sigma_q^{\mathrm{tr}}+\epsilon},
    \label{eq:score_standardization}
\end{equation}
where
\begin{equation}
    \mu_{S_t}^{\mathrm{tr}}=\operatorname{mean}_{u} S_t^{\mathrm{tr}}(u), \quad
    \sigma_b^{\mathrm{tr}}=\operatorname{std}_{u} S_t^{\mathrm{tr}}(u),
\end{equation}
and $\mu_g^{\mathrm{tr}},\sigma_g^{\mathrm{tr}},\mu_q^{\mathrm{tr}},\sigma_q^{\mathrm{tr}}$ are defined analogously.

The final PAI score is a weighted sum:
\begin{equation}
    S_{\mathrm{PAI}}(t)
    =
    w_b z_{euc}(t)+\lambda_g z_{magG}(t)+\lambda_q z_{T_2}(t).
    \label{eq:operational_score}
\end{equation}
The fixed configuration $\Lambda_{\mathrm{fixed}}=(w_b,\lambda_g,\lambda_q)$ is selected on TSB-AD and then evaluated on \textsc{TAB} UV without TAB-specific tuning. Numeric values and robustness analyses are reported in Appendix~\ref{app:weight_robustness}.

\begin{table}[!ht]
\centering
\caption{\textbf{Main results on TSB-AD-U-Eva.} We report univariate time-series anomaly detection performance across classical, neural, and representation-based methods. For each metric, the best and second-best values are shown in \textbf{bold} and \underline{underlined}, respectively. All scores are averaged over five runs.}
\label{tab:finalresult_tsbadu}

\vspace{-1ex}
\renewcommand{\arraystretch}{1.00}

\begin{adjustbox}{max width=\textwidth}
\begin{tabular}{@{}>{\centering\arraybackslash}m{0.55cm}@{} | l c c c c c c@{}}
\toprule
\multicolumn{2}{c}{} 
  & \multicolumn{3}{c}{Range-Wise Measure $\uparrow$} 
  & \multicolumn{3}{c}{Point-Wise Measure $\uparrow$} \\
\cmidrule(lr){3-5}\cmidrule(lr){6-8}
\multicolumn{1}{c}{} & \textbf{\shortstack[b]{Method}} 
  & \textbf{VUS-PR} & VUS-ROC & Range-F1 
  & AUC-PR & AUC-ROC & Point-F1 \\
\midrule

\multirow{5}{*}{\rotatebox[origin=c]{90}{\footnotescriptsize\textbf{{Stat\&ML}}}}
 & (Sub)-PCA~\cite{charu2017pca}       & 0.42 & 0.76 & 0.41 & 0.37 & 0.71 & 0.42 \\
 & KShapeAD~\cite{papa2017kshape2}     & 0.40 & 0.76 & 0.40 & 0.35 & 0.74 & 0.39 \\
& \texttt{magG} $g(t)$      & 0.34 & 0.76 & 0.61 & 0.34 & 0.70 & 0.38 \\
& T2 $q(t)$                & 0.36 & 0.71 & 0.44 & 0.33 & 0.66 & 0.36 \\
& \texttt{magG}+T2       & 0.38 & 0.78 & 0.56 & 0.36 & 0.72 & 0.40 \\
 
\midrule
\multirow{6}{*}{\rotatebox[origin=c]{90}{\footnotescriptsize\textbf{{NN}}}}
 & DeepAnT~\cite{munir2018cnn}        & 0.34 & 0.79 & 0.35 & 0.33 & 0.71 & 0.38 \\
 & USAD~\cite{audibert2020usad}       & 0.36 & 0.71 & 0.40 & 0.32 & 0.66 & 0.37\\
 & TimesNet~\cite{wu2023timesnet}     & 0.26 & 0.72 & 0.21 & 0.18 & 0.61 & 0.24 \\
 & FITS~\cite{xu2024fits}             & 0.26 & 0.73 & 0.20 & 0.17 & 0.61 & 0.23 \\
& DADA~\cite{shentu2025towards}         & 0.31 & 0.77 & 0.31 & 0.29 & 0.71 & 0.38 \\
 
 & KAN-AD~\cite{zhou2025kanad}      & 0.43 & 0.82 & 0.43 & 0.41 & 0.80 & 0.44\\
 
\midrule
\multirow{10}{*}{\rotatebox[origin=c]{90}{\footnotescriptsize\textbf{{Representation}}}}
& TS2Vec [AAAI'22]~\cite{yue2022ts2vec}  & 0.19 & 0.62 & 0.60 & 0.19 & 0.54 & 0.25  \\
& \cc{\textbf{+ PAI}} & \cc{0.47} & \cc{0.87} & \cc{0.67} & \cc{0.45} & \cc{0.83} & \cc{0.48} \\
\cline{2-8}
& DCdetector [KDD'23]~\cite{yang2023dcdetector}  &  0.10 & 0.55 & 0.13 & 0.06 & 0.48 & 0.11  \\
& \cc{\textbf{+ PAI}} & \cc{0.39} & \cc{0.81} & \cc{0.66} & \cc{0.38} & \cc{0.76} & \cc{0.43} \\
\cline{2-8}
& TSPulse (ZS) [ICLR'26]~\cite{ekambaram2026tspulse}  & 0.41 & 0.79 & 0.50 & 0.32 & 0.74 & 0.39  \\
& \cc{\textbf{+ PAI}}  & \cc{0.48} & \cc{0.85} & \cc{0.61} & \cc{0.43} & \cc{0.81} & \cc{0.48}\\
\cline{2-8}
& TSPulse (FT) [ICLR'26]~\cite{ekambaram2026tspulse}  & 0.54 & 0.87 & 0.67 & 0.48 & 0.84 & 0.54 \\
& \cc{\textbf{+ PAI}} & \cc{\underline{0.58}} & \cc{\underline{0.90}} & \cc{\underline{0.71}} & \cc{\underline{0.52}} & \cc{\underline{0.87}} & \cc{\textbf{0.58}}\\
\cline{2-8}
& PaAno [ICLR'26]~\cite{park2026paano}  & 0.52 & 0.89 & 0.48 & 0.46 & 0.86 & 0.51 \\
& \cc{\textbf{+ PAI}} & \cc{\textbf{0.58}} & \cc{\textbf{0.93}} & \cc{\textbf{0.75}} & \cc{\textbf{0.55}} & \cc{\textbf{0.90}} & \cc{\underline{0.58}}   \\
\bottomrule
\end{tabular}
\end{adjustbox}
\vspace{-10pt}
\end{table}

\section{Experiments} 
\label{sec:experiments}

\subsection{Experimental setup}
\label{sec:setup}

\textbf{Datasets.}
We use TSB-AD-U-Eva~\citep{liu2024the} as the primary benchmark for main comparisons, ablations, and diagnostics. This benchmark contains 350 univariate time-series instances from TSB-AD-U. To evaluate cross-corpus transfer, we further test on \textsc{TAB} UV~\citep{qiu2025tab}, a larger univariate benchmark with 1,635 time-series instances. All PAI fusion parameters are selected under the TSB-AD-U-Eva protocol and reused on \textsc{TAB} UV without additional tuning.

\textbf{Compared methods.}
We compare against three groups of baselines: classical statistical and machine-learning methods, neural time-series anomaly detection methods, and representation-based methods. The first group includes (Sub)-PCA~\citep{charu2017pca}, KShapeAD~\citep{papa2017kshape2}, and the raw-scale scores \texttt{magG}, T2, and \texttt{magG}+T2. The neural baselines include DeepAnT~\citep{munir2018cnn}, USAD~\citep{audibert2020usad}, TimesNet~\citep{wu2023timesnet}, FITS~\citep{xu2024fits}, DADA~\citep{shentu2025towards}, and KAN-AD~\citep{zhou2025kanad}. For representation-based methods, we evaluate TS2Vec~\citep{yue2022ts2vec}, DCdetector~\citep{yang2023dcdetector}, TSPulse~\citep{ekambaram2026tspulse}, and PaAno~\citep{park2026paano}, and report each base score together with its PAI-augmented variant.

\textbf{PAI variants.}
For TS2Vec, DCdetector, and PaAno, the base score is the Euclidean representation-bank score defined in Equation~\ref{eq:timestamp_bank_score}. The corresponding ``+ PAI'' variant augments this score with the two training-calibrated raw-amplitude terms, \texttt{magG} and T2, using the fixed fusion rule in Equation~\ref{eq:operational_score}. For TSPulse, we evaluate both zero-shot and fine-tuned settings. Since its native prediction/reconstruction head provides a stronger anomaly score than hidden-state bank distance, the TSPulse ``+ PAI'' variant keeps the native head score as the base score and applies the same raw-amplitude augmentation. The same fusion weights are used across all PAI variants.

\textbf{Metrics.}
Following TSB-AD~\citep{liu2024the}, we report VUS-PR, VUS-ROC, Range-F1, AUC-PR, AUC-ROC, and Point-F1. VUS-PR is the primary metric because anomalies are rare and range-level precision--recall behavior is central to this setting. Scores presented in the table are averaged over five runs.

\subsection{Main Results on TSB-AD-U-Eva}
\label{sec:main_results}

Table~\ref{tab:finalresult_tsbadu} reports the main comparison on TSB-AD-U-Eva. The evaluated methods are grouped into classical statistical or machine-learning baselines (\textbf{Stat\&ML}), neural anomaly detection baselines (\textbf{NN}), and representation-based methods (\textbf{Representation}). PAI improves every reported metric for all five representation-based configurations in Table~\ref{tab:finalresult_tsbadu}, giving 30/30 positive configuration-metric deltas. On the primary VUS-PR metric, TS2Vec improves from $0.19$ to $0.47$, DCdetector from $0.10$ to $0.39$, TSPulse$_{\mathrm{ZS}}$ from $0.41$ to $0.48$, TSPulse$_{\mathrm{FT}}$ from $0.54$ to $0.58$, and PaAno from $0.52$ to $0.58$. PaAno + PAI obtains the best VUS-PR, VUS-ROC, Range-F1, AUC-PR, and AUC-ROC, while TSPulse$_{\mathrm{FT}}$ + PAI obtains the best Point-F1. We report confidence intervals for each performance gain in Appendix~\ref{app:paired_ci}.

The raw-amplitude scores are also informative by themselves: \texttt{magG}+T2 reaches $0.38$ VUS-PR without any learned representation. However, raw amplitude alone does not explain the strongest results. TS2Vec + PAI, TSPulse$_{\mathrm{FT}}$ + PAI, and PaAno + PAI all exceed the raw-only score, indicating that the best-performing variants combine learned temporal representations with explicit raw-amplitude evidence. For TS2Vec and DCdetector, the reported base rows use the unified representation-bank scoring protocol described in Section~\ref{sec:problem_setting}, rather than the original method-specific scoring rules. This choice allows us to evaluate bank-compatible encoders under a common scoring protocol and isolate the contribution of PAI. Results using the original scoring rules are reported in Appendix~\ref{app:original_scorers}. Parameter counts and run times are reported in Appendix~\ref{app:runtime_params}.

\subsection{Score-Component and Normalization Ablation Study}
\label{sec:diagnostic_results}
\begin{wraptable}{ht}{0.5\textwidth}
    \vspace{-18pt}
\caption{\textbf{Score-component ablation on TSB-AD-U-Eva}, reported as VUS-PR.}
    \vspace{-10pt}
    \label{tab:eva350_component_ablation_compact}
    \begin{center}
    \resizebox{1.0\hsize}{!}{
    \begin{tabular}{lcccc}
    \toprule
Method & Base  & Euclidean  & Raw  & + PAI \\
& score & bank & only & \\
\midrule
DCdetector & 0.10 & 0.24 & 0.38 & \textbf{0.39} \\
TS2Vec & 0.19 & 0.47 & 0.38 & \textbf{0.47} \\
TSPulse$_{\mathrm{ZS}}$ & 0.41 & 0.11 & 0.38 & \textbf{0.48} \\
PaAno & 0.51 & 0.52 & 0.38 & \textbf{0.58} \\
 \bottomrule
    \end{tabular}}
    \end{center}
    \vspace{-15pt}
\end{wraptable}

We analyze how each PAI component contributes to the overall gains. PAI improves representation-based scoring in two ways: it replaces cosine-based bank scoring with Euclidean bank scoring, and it adds explicit raw-amplitude scores. When the embedding norm already contains useful amplitude information, Euclidean scoring should provide most of the gain. When the representation score is largely scale-invariant, the raw-amplitude terms should contribute more by restoring information missing from the learned representation.

Table~\ref{tab:eva350_component_ablation_compact} shows that different methods expose amplitude information differently. TS2Vec~\cite{yue2022ts2vec} is the clearest radial-information case: Euclidean bank scoring reaches $0.47$ VUS-PR, nearly matching the full PAI score. This suggests that, under file-wise normalization, amplitude-related deviations are largely preserved in the TS2Vec embedding norm. DCdetector~\cite{yang2023dcdetector} shows a mixed pattern: Euclidean scoring improves VUS-PR from $0.10$ to $0.24$, while full PAI further increases it to $0.39$, indicating that explicit raw-amplitude scores remain important. PaAno~\cite{park2026paano} is closer to amplitude-invariant behavior: Euclidean scoring provides only a small gain over cosine scoring, whereas adding raw-amplitude scores raises VUS-PR from $0.52$ to $0.58$. A more detailed component ablation is provided in Appendix~\ref{app:score_components} and~\ref{app:tab_uv_ci}.

TSPulse~\citep{ekambaram2026tspulse} serves as a useful control. Its native zero-shot head achieves $0.41$ VUS-PR, while Euclidean bank scoring over hidden states reaches only $0.11$. We therefore keep the native TSPulse score and augment it with the same \texttt{magG} and \texttt{T2} terms, increasing zero-shot VUS-PR to $0.48$.

\subsection{PAI Gains Across Anomaly Types}
\label{sec:types}

\begin{wrapfigure}{r}{0.48\linewidth}
    \vspace{-12pt}
    \centering
    \includegraphics[width=1.0\linewidth]{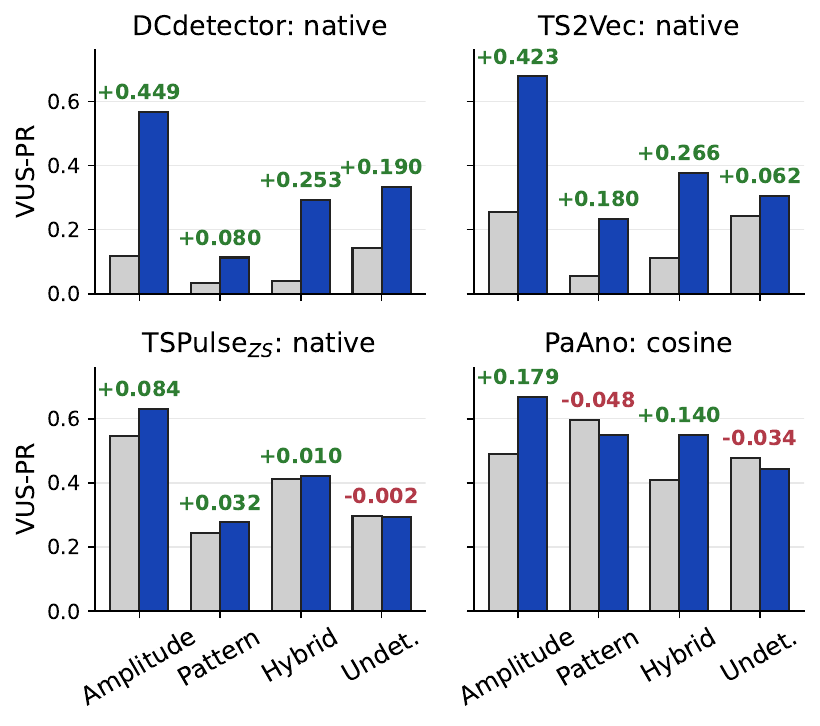}
    \vspace{-13pt}
    \caption{
    \textbf{VUS-PR by TSB-AD anomaly type.}
    }
    \label{fig:eva350_per_type_breakdown}
    \vspace{-15pt}
\end{wrapfigure}

If PAI preserves raw-amplitude information, its should ehance the performance of the base methods on anomalies involving amplitude or level changes. Figure~\ref{fig:eva350_per_type_breakdown} reports VUS-PR by anomaly type on TSB-AD-U-Eva~\citep{liu2024the}.

The results are consistent with the proposed mechanism. All four methods improve on amplitude anomalies, with the largest gains for DCdetector and TS2Vec. Hybrid anomalies also improve across all methods, suggesting that raw-amplitude information remains useful when amplitude changes occur simultaneously with temporal or shape changes. The same breakdown also shows the limitation of PAI. For PaAno, PAI improves amplitude and hybrid anomalies but slightly decreases performance on pattern anomalies and undetermined cases. This tradeoff is expected: explicit raw-amplitude terms help when amplitude or level is discriminative, but they can perturb the ranking when shape or temporal pattern is the dominant signal.

\subsection{Cross-Corpus Transfer to \textsc{TAB} UV}
\label{sec:tab_uv}

\begin{wraptable}{ht}{0.5\textwidth}
    \vspace{-18pt}
\caption{\textbf{Cross-corpus transfer to \textsc{TAB} UV.}}
    \vspace{-10pt}
    \label{tab:tab_uv_representation_main}
    \begin{center}
    \resizebox{1.0\hsize}{!}{
    \begin{tabular}{llrr}
    \toprule
Method & Variant & VUS-PR & AUC-PR \\
\midrule
\multirow{2}{*}{TS2Vec}
 & cosine & 0.21 & 0.17 \\
 & + PAI & \textbf{0.25} & \textbf{0.26} \\
\midrule
\multirow{2}{*}{DCdetector}
 & cosine & 0.12 & 0.09 \\
 & + PAI & \textbf{0.26} & \textbf{0.25} \\
\midrule
\multirow{2}{*}{TSPulse$_{\mathrm{ZS}}$}
 & native & 0.22 & 0.14 \\
 & native + PAI & \textbf{0.26} & \textbf{0.23} \\
 \midrule
\multirow{2}{*}{PaAno}
 & cosine & 0.20 & 0.20 \\
 & native + PAI & \textbf{0.26} & \textbf{0.29} \\
 \bottomrule
    \end{tabular}}
    \end{center}
    \vspace{-18pt}
\end{wraptable}

We then evaluate whether the same score-fusion recipe transfers to a different univariate benchmark. Table~\ref{tab:tab_uv_representation_main} reports \textsc{TAB} UV results for the protocol-aligned methods: TS2Vec, DCdetector, TSPulse$_{\mathrm{ZS}}$, and PaAno. The PAI parameters are fixed from TSB-AD-U-Eva.

Although absolute scores are lower on \textsc{TAB} UV, the improvement direction is preserved. PAI improves TS2Vec from $0.209$ to $0.251$ VUS-PR, DCdetector from $0.120$ to $0.258$, and TSPulse$_{\mathrm{ZS}}$ from $0.223$ to $0.262$. AUC-PR also improves for all methods. These results support PAI as a transferable score-augmentation strategy. Weight-grid sensitivity results are reported in Appendix~\ref{app:weight_robustness}, showing that the fixed recipe is not an isolated TSB-specific optimum.

\begin{figure}[t]
\centering
\includegraphics[width=\textwidth]{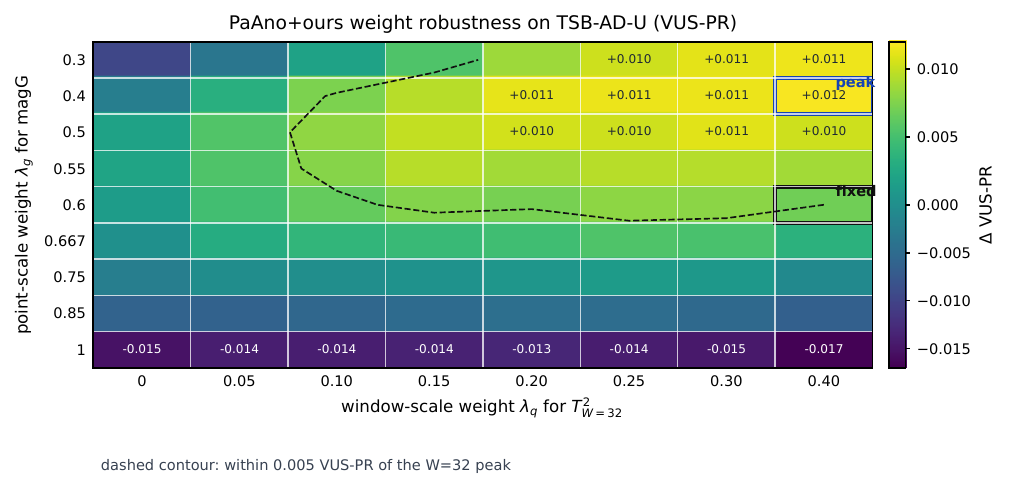}
\caption{
\textbf{Weight robustness of PaAno + PAI on TSB-AD-U-Eva.}
The heat map shows the full weight grid for the fusion rule
$z_{euc}+\lambda_g z_{mag}+\lambda_q z_{T_2}$ at $W=32$, reported as the
five-seed mean $\Delta$ VUS-PR relative to the PaAno base score.
The selected fixed configuration lies within a broad positive plateau rather
than at an isolated optimum.
}
\label{fig:a30_weight_sensitivity}
\vspace{-15pt}
\end{figure}

\subsection{Robustness and Computational Cost}
\label{sec:robustness_cost}

We test whether the gain of PAI depends on a narrow choice of fusion weights. Figure~\ref{fig:a30_weight_sensitivity} shows the full weight grid for PaAno + PAI on TSB-AD-U-Eva. The performance surface forms a broad positive plateau, indicating that the selected fixed configuration is not a finely tuned optimum but lies within a stable region. The final weights we used in the experiments is $w_b=1$, $\lambda_g=0.6$ and $\lambda_q=0.4$ instead of the peak to trade off the performance gain on TAB dataset. The complete grid is reported in Appendix~\ref{app:weight_grid}. The additional raw-amplitude terms introduce negligible computational overhead. 

We report parameter counts and runtimes for the benchmark methods in Appendix~\ref{app:runtime_params}. The key result is that PAI adds negligible overhead: on a representative TSB-AD-U-Eva time series of length 4{,}031, computing \texttt{magG}+\texttt{T2} takes only $0.003$ seconds and requires no GPU memory, whereas training a TS2Vec encoder takes $18.7$ seconds and 121\,MB. Thus, PAI improves raw-amplitude sensitivity at the scoring stage without increasing the training cost of the base representation method.

\subsection{Limitations}
\label{sec:experiment_limitations}

PAI consistently improves magnitude and hybrid anomalies, but the anomaly-type breakdown in Section \ref{sec:types} shows that PaAno + PAI slightly decreases performance on pattern and undetermined anomalies. Thus, PAI is best viewed as a score-fusion method for incorporating complementary raw-amplitude evidence, rather than a universal replacement for representation-based scoring. When anomalies are defined mainly by temporal shape instead of amplitude or level, fixed raw-amplitude terms can perturb the ranking. In addition, our evaluation is limited to univariate time series. The raw-signal scores \texttt{magG} and T2 are computed per series and do not model cross-channel dependencies, so multivariate extension requires additional design.

\section{Conclusion}
\label{sec:conclusion}

This paper identifies a critical limitation of representation-based univariate TSAD methods: their final anomaly scores can be largely amplitude-agnostic, causing amplitude-related anomalies such as spikes, level shifts, and intensity changes to be under-scored. To address this issue, we introduced PAI, a lightweight score-level augmentation that requires no changes to the base encoder or training procedure. PAI first uses a cosine-versus-Euclidean diagnostic to assess whether amplitude information is preserved in the learned representation, and then combines a Euclidean representation score with two raw-amplitude scores, \texttt{magG} and \texttt{T2}. Experiments on TSB-AD-U-Eva and \textsc{TAB} UV show that PAI consistently improves all four evaluated representation-based methods across every reported metric, achieving average VUS-PR gains of 98.4\% and 36.8\%, respectively. Among all evaluated combinations, PaAno + PAI achieves the best overall performance, outperforming the state-of-the-art method by 15\%. The gains are especially clear on time series dominated by amplitude anomalies, where PAI substantially improves all evaluated representation-based methods. These results show that explicit amplitude-aware scoring is a simple and effective correction for more robust representation-based TSAD.

{
    \small
    \bibliography{main}
    \bibliographystyle{unsrt}
}


\appendix

\section{Additional Experimental Details}
\label{app:experiments}

\subsection{Paired Uncertainty on TSB-AD-U-Eva}
\label{app:paired_ci}

Table~\ref{tab:ci_paired} reports paired bootstrap confidence intervals for the PAI gain over the corresponding base score on TSB-AD-U-Eva. The bootstrap resamples time-series instances and preserves the pairing between the base score and the PAI score. The intervals exclude zero for all four primary representation methods, supporting the main-text claim that the improvements are not driven by a small set of sampled instances.

\begin{table}[ht]
\centering
\small
\caption{Paired bootstrap 95\% confidence intervals for $\Delta$ (PAI score minus base score) on TSB-AD-U-Eva. We use $n_{\mathrm{boot}}=2000$ resamples with seed $42$, paired over the common set of time-series instances.}
\label{tab:ci_paired}
\begin{tabular}{lcc}
\toprule
Method & $\Delta$ VUS-PR $[$95\% CI$]$ & $\Delta$ AUC-PR $[$95\% CI$]$ \\
\midrule
DCdetector  & $+0.293$ $[+0.259,\,+0.328]$ & $+0.317$ $[+0.282,\,+0.353]$ \\
TS2Vec      & $+0.279$ $[+0.240,\,+0.316]$ & $+0.259$ $[+0.218,\,+0.297]$ \\
TSPulse$_{\mathrm{ZS}}$ & $+0.048$ $[+0.031,\,+0.068]$ & $+0.054$ $[+0.038,\,+0.071]$ \\
PaAno       & $+0.075$ $[+0.049,\,+0.103]$ & $+0.092$ $[+0.064,\,+0.122]$ \\
\bottomrule
\end{tabular}
\end{table}

\subsection{Reference Scores and Bank-Score Variants}
\label{app:original_scorers}

The main table reports each representation method together with its PAI score. For bank-compatible methods, PAI uses Euclidean representation-bank scoring as the representation term; for TSPulse, PAI keeps the native prediction/reconstruction score because hidden-state bank scoring is not competitive. Table~\ref{tab:reference_and_bank_scores} separates these score pathways on TSB-AD-U-Eva. This distinction is important: a weak reference score does not imply that the encoder lacks useful information, since Euclidean bank scoring may expose amplitude-related information through the representation norm.

\begin{table}[ht]
\centering
\small
\caption{Reference scores and bank-score variants on TSB-AD-U-Eva, reported as VUS-PR. The reference score is the base row reported in the main comparison. Euclidean bank scoring is used as the representation term for bank-compatible PAI variants. For TSPulse, encoder-bank scoring is diagnostic only; the PAI score keeps the native head.}
\label{tab:reference_and_bank_scores}
\begin{adjustbox}{max width=\textwidth}
\begin{tabular}{lcccl}
\toprule
Method & Reference score & Cosine bank & Euclidean bank & PAI score \\
\midrule
DCdetector & 0.097 & 0.165 & 0.236 & \textbf{0.390} \\
TS2Vec & 0.193 & 0.384 & 0.470 & \textbf{0.473} \\
TSPulse$_{\mathrm{ZS}}$ & 0.411 & -- & 0.107 & \textbf{0.484} \\
PaAno & 0.509 & 0.509 & 0.521 & \textbf{0.584} \\
\bottomrule
\end{tabular}
\end{adjustbox}
\end{table}

\section{Full Score-Term Ablation}
\label{app:score_components}

Table~\ref{tab:eva350_component_ablation} expands the component ablation summarized in the main text. The table isolates Euclidean representation-bank scoring, the two raw-scale scores, and their combinations. TS2Vec already obtains most of its gain from Euclidean representation distance, indicating that its representation norm contains useful amplitude-related information under the current preprocessing. DCdetector improves when all terms are combined, indicating that explicit raw-scale scores remain complementary. For TSPulse, encoder-bank distance is weak even after raw-scale fusion, which is why the main experiments keep the native head score for TSPulse + PAI.

\begin{table}[t]
\centering
\scriptsize
\caption{Component ablation on TSB-AD-U-Eva. Rows isolate the Euclidean representation-bank term, the two raw-scale scores, and their pairwise/full fusions. The TSPulse rows use encoder-bank distance for the representation term; the native prediction/reconstruction head is not included in this table.}
\label{tab:eva350_component_ablation}
\begin{adjustbox}{max width=\textwidth}
\begin{tabular}{llrrrr}
\toprule
Method & Score term(s) & VUS-PR & VUS-ROC & AUC-PR & AUC-ROC \\
\midrule
\multirow{7}{*}{TS2Vec}
 & Euclidean rep. only & 0.470 & 0.869 & 0.436 & 0.823 \\
 & \texttt{magG} only & 0.342 & 0.760 & 0.337 & 0.697 \\
 & T2 only & 0.355 & 0.713 & 0.328 & 0.662 \\
 & Euclidean + \texttt{magG} & 0.469 & 0.869 & 0.440 & 0.826 \\
 & Euclidean + T2 & \textbf{0.475} & \textbf{0.871} & 0.443 & 0.829 \\
 & \texttt{magG} + T2 & 0.375 & 0.777 & 0.364 & 0.720 \\
 & Euclidean + \texttt{magG} + T2 & 0.473 & 0.869 & \textbf{0.447} & \textbf{0.829} \\
\midrule
\multirow{7}{*}{DCdetector}
 & Euclidean rep. only & 0.236 & 0.715 & 0.207 & 0.661 \\
 & \texttt{magG} only & 0.342 & 0.760 & 0.337 & 0.697 \\
 & T2 only & 0.355 & 0.713 & 0.328 & 0.662 \\
 & Euclidean + \texttt{magG} & 0.318 & 0.784 & 0.302 & 0.734 \\
 & Euclidean + T2 & 0.329 & 0.791 & 0.298 & 0.741 \\
 & \texttt{magG} + T2 & 0.375 & 0.777 & 0.364 & 0.720 \\
 & Euclidean + \texttt{magG} + T2 & \textbf{0.390} & \textbf{0.810} & \textbf{0.375} & \textbf{0.762} \\
\midrule
\multirow{7}{*}{TSPulse$_{\mathrm{ZS}}$ encoder}
 & Euclidean rep. only & 0.107 & 0.586 & 0.068 & 0.519 \\
 & \texttt{magG} only & 0.342 & 0.760 & 0.337 & 0.697 \\
 & T2 only & 0.355 & 0.713 & 0.328 & 0.662 \\
 & Euclidean + \texttt{magG} & 0.161 & 0.683 & 0.124 & 0.615 \\
 & Euclidean + T2 & 0.188 & 0.678 & 0.147 & 0.620 \\
 & \texttt{magG} + T2 & \textbf{0.375} & \textbf{0.777} & \textbf{0.364} & \textbf{0.720} \\
 & Euclidean + \texttt{magG} + T2 & 0.237 & 0.716 & 0.199 & 0.652 \\
\bottomrule
\end{tabular}
\end{adjustbox}
\end{table}

Table~\ref{tab:eva350_ts2vec_revin_ablation} reports the input-normalization ablation used to interpret the cosine--Euclidean diagnostic. Without input normalization, Euclidean scoring improves substantially over cosine scoring, and adding raw-scale terms on top of Euclidean scoring changes little. With input normalization, the Euclidean-over-cosine gap becomes smaller, while the raw-scale terms become more complementary. This supports the mechanism that normalization changes where amplitude information is available: either in the representation norm or in explicit raw-signal scores.

\begin{table}[t]
\centering
\small
\caption{TS2Vec with and without input normalization on TSB-AD-U-Eva. Input normalization reduces the Euclidean-minus-cosine gap, but it also lowers the pure cosine representation score. The ablation therefore supports the normalization mechanism with this caveat.}
\label{tab:eva350_ts2vec_revin_ablation}
\begin{adjustbox}{max width=\textwidth}
\begin{tabular}{llrrrr}
\toprule
Encoder & Variant & VUS-PR & VUS-ROC & AUC-PR & AUC-ROC \\
\midrule
\multirow{5}{*}{Vanilla TS2Vec}
 & native & 0.193 & 0.620 & 0.189 & 0.539 \\
 & cosine bank & 0.384 & 0.847 & 0.329 & 0.791 \\
 & Euclidean bank & 0.470 & 0.869 & 0.436 & 0.823 \\
 & cosine bank + raw-scale & 0.450 & 0.866 & 0.403 & 0.824 \\
 & Euclidean bank + raw-scale & \textbf{0.473} & \textbf{0.869} & \textbf{0.447} & \textbf{0.829} \\
\midrule
\multirow{5}{*}{TS2Vec + input normalization}
 & native & 0.193 & 0.620 & 0.189 & 0.539 \\
 & cosine bank & 0.323 & 0.792 & 0.264 & 0.731 \\
 & Euclidean bank & 0.356 & 0.801 & 0.322 & 0.751 \\
 & cosine bank + raw-scale & 0.457 & 0.872 & 0.411 & 0.824 \\
 & Euclidean bank + raw-scale & 0.445 & 0.853 & 0.428 & 0.804 \\
\bottomrule
\end{tabular}
\end{adjustbox}
\end{table}

\section{Cross-Corpus Transfer to TAB UV}
\label{app:tab_uv_ci}

Table~\ref{tab:tab_uv_cross_corpus} reports the protocol-aligned \textsc{TAB} UV transfer results used in the main text. PAI uses the same score-fusion parameters selected on TSB-AD-U-Eva, with no TAB-specific tuning. We include only the methods evaluated under the same TAB-side protocol: TS2Vec, DCdetector, and TSPulse$_{\mathrm{ZS}}$.

\begin{table}[t]
\centering
\small
\caption{Protocol-aligned cross-corpus results on \textsc{TAB} UV. PAI uses the fixed score-fusion parameters selected on TSB-AD-U-Eva.}
\label{tab:tab_uv_cross_corpus}
\begin{adjustbox}{max width=\textwidth}
\begin{tabular}{llrrrrrr}
\toprule
Method & Variant & VUS-PR & VUS-ROC & Range-F1 & AUC-PR & AUC-ROC & Point-F1 \\
\midrule
\multirow{2}{*}{TS2Vec}
 & cosine bank & 0.209 & 0.681 & 0.330 & 0.168 & 0.631 & 0.232 \\
 & + PAI  & \textbf{0.251} & \textbf{0.715} & \textbf{0.415} & \textbf{0.263} & \textbf{0.687} & \textbf{0.316} \\
\midrule
\multirow{2}{*}{DCdetector}
 & cosine bank & 0.120 & 0.707 & 0.257 & 0.091 & 0.614 & 0.138 \\
 & + PAI  & \textbf{0.258} & \textbf{0.774} & \textbf{0.439} & \textbf{0.251} & \textbf{0.741} & \textbf{0.311} \\
\midrule
\multirow{2}{*}{TSPulse$_{\mathrm{ZS}}$}
 & native & 0.223 & 0.641 & 0.230 & 0.137 & 0.613 & 0.196 \\
 & native + PAI & \textbf{0.262} & \textbf{0.699} & \textbf{0.337} & \textbf{0.229} & \textbf{0.668} & \textbf{0.292} \\
\bottomrule
\end{tabular}
\end{adjustbox}
\end{table}

Table~\ref{tab:tab_uv_representation_full} separates cosine distance, Euclidean distance, raw-scale fusion, and TSPulse native-head fusion on \textsc{TAB} UV. These results are mechanism support for the transfer analysis rather than a separate leaderboard claim.

\begin{table}[t]
\centering
\scriptsize
\caption{Full protocol-aligned representation-method comparison on \textsc{TAB} UV. The table separates cosine distance, Euclidean distance, raw-scale fusion, and TSPulse native-head fusion. Non-protocol-aligned PaAno-family TAB results are omitted.}
\label{tab:tab_uv_representation_full}
\begin{adjustbox}{max width=\textwidth}
\begin{tabular}{llrrrrrr}
\toprule
Method & Variant & VUS-PR & VUS-ROC & Range-F1 & AUC-PR & AUC-ROC & Point-F1 \\
\midrule
\multirow{4}{*}{TS2Vec}
 & cosine bank & 0.209 & 0.681 & 0.330 & 0.168 & 0.631 & 0.232 \\
 & Euclidean bank & 0.245 & 0.687 & 0.404 & 0.254 & 0.656 & 0.302 \\
 & cosine bank + raw-scale & 0.245 & 0.715 & 0.398 & 0.231 & 0.684 & 0.297 \\
 & Euclidean bank + raw-scale & \textbf{0.251} & \textbf{0.715} & \textbf{0.415} & \textbf{0.263} & \textbf{0.687} & \textbf{0.316} \\
\midrule
\multirow{4}{*}{DCdetector}
 & cosine bank & 0.120 & 0.707 & 0.257 & 0.091 & 0.614 & 0.138 \\
 & Euclidean bank & 0.216 & 0.737 & 0.377 & 0.177 & 0.694 & 0.238 \\
 & cosine bank + raw-scale & 0.214 & 0.754 & 0.387 & 0.220 & 0.710 & 0.270 \\
 & Euclidean bank + raw-scale & \textbf{0.258} & \textbf{0.774} & \textbf{0.439} & \textbf{0.251} & \textbf{0.741} & \textbf{0.311} \\
\midrule
\multirow{6}{*}{TSPulse$_{\mathrm{ZS}}$}
 & native & 0.223 & 0.641 & 0.230 & 0.137 & 0.613 & 0.196 \\
 & cosine bank & 0.048 & 0.560 & 0.102 & 0.028 & 0.487 & 0.062 \\
 & Euclidean bank & 0.048 & 0.566 & 0.092 & 0.027 & 0.480 & 0.061 \\
 & cosine bank + raw-scale & 0.101 & 0.661 & 0.201 & 0.096 & 0.608 & 0.146 \\
 & Euclidean bank + raw-scale & 0.106 & 0.662 & 0.204 & 0.100 & 0.609 & 0.160 \\
 & native + raw-scale & \textbf{0.262} & \textbf{0.699} & \textbf{0.337} & \textbf{0.229} & \textbf{0.668} & \textbf{0.292} \\
\bottomrule
\end{tabular}
\end{adjustbox}
\end{table}

\section{Weight Robustness and Computational Cost}
\label{app:weight_robustness}

\subsection{Fusion-Weight Robustness}
\label{app:weight_grid}

The weight checks are included to rule out the interpretation that PAI depends on a narrowly tuned set of raw-scale weights. The main comparison uses a fixed score-fusion recipe; the appendix shows that nearby weights produce the same qualitative behavior.

Table~\ref{tab:eva350_weight_grid_full} reports the full TSB-AD-U-Eva grid for TS2Vec, DCdetector, and the TSPulse encoder-bank diagnostic. TS2Vec and DCdetector have broad positive regions rather than isolated optima. The TSPulse encoder rows reinforce the native-head choice in the main text: when TSPulse is forced into encoder-bank scoring, raw-scale fusion does not recover native-head performance.

\subsection{Parameter Counts and Run Time}
\label{app:runtime_params}

Table~\ref{tab:runtime_params} reports parameter counts and run times for the benchmark methods. The PAI score does not introduce learned parameters. Its raw-scale terms are computed directly from the input signal and add approximately $0.003$s on a representative TSB-AD-U-Eva time-series instance of length 4{,}031.

\begin{table}[ht]
\centering
\scriptsize
\caption{Parameter counts and run times for the benchmark methods. PAI score-fusion rows inherit the learned parameters of the base method; the raw-scale terms have no learned parameters. A dash denotes nonparametric methods or unavailable counts in the source table.}
\label{tab:runtime_params}
\begin{adjustbox}{max width=\textwidth}
\begin{tabular}{lrr}
\toprule
Method & \# Params & Run Time \\
\midrule
\multicolumn{3}{l}{\textbf{Statistical and machine-learning methods}} \\
(Sub)-PCA & -- & 1.5s \\
KShapeAD & -- & 8.0s \\
DLinear & $<0.1$M & 2.9s \\
NLinear & $<0.1$M & 5.8s \\
\midrule
\multicolumn{3}{l}{\textbf{Neural methods}} \\
DeepAnT & $<0.1$M & 2.0s \\
USAD & $<0.1$M & 1.7s \\
TimesNet & $<0.1$M & 11.2s \\
FITS & $<0.1$M & 3.1s \\
DADA & $1.84$M & 0.8s \\
KAN-AD & $<0.1$M & 12.1s \\
\midrule
\multicolumn{3}{l}{\textbf{Transformer methods}} \\
AnomalyTransformer & $4.8$M & 48.9s \\
DCdetector & $0.9$M & 5.8s \\
Lag-Llama & $2.5$M & 1220.8s \\
OFA & $81.9$M & 171.1s \\
PatchTST & $0.5$M & 26.3s \\
iTransformer & $0.6$M & 9.8s \\
MOMENT (FT) & $109.6$M & 43.6s \\
MOMENT (ZS) & $109.6$M & 42.9s \\
TimesFM & $203.5$M & 83.8s \\
\midrule
\multicolumn{3}{l}{\textbf{Representation-based methods and PAI terms}} \\
DCdetector & $0.9$M & 5.8s \\
PaAno & $0.3$M & 6.9s \\
\texttt{magG}+T2 score terms & $0$ & $+0.003$s \\
\bottomrule
\end{tabular}
\end{adjustbox}
\end{table}

\begin{table}[t]
\centering
\scriptsize
\caption{Weight-grid robustness on TSB-AD-U-Eva. Each row evaluates $w_e z(e)+\lambda_g z(\texttt{magG})+\lambda_q z(\mathrm{T2})$ on the common time-series instances. Bold marks the best value in each metric within a method block.}
\label{tab:eva350_weight_grid_full}
\begin{adjustbox}{max width=\textwidth}
\begin{tabular}{lrrrrrrrr}
\toprule
Method & $w_e$ & $\lambda_g$ & $\lambda_q$ & $n$ & VUS-PR & VUS-ROC & AUC-PR & AUC-ROC \\
\midrule
\multirow{16}{*}{DCdetector}
 & 0.0 & 0.0 & 1.0 & 349 & 0.355 & 0.713 & 0.328 & 0.662 \\
 & 0.0 & 1.0 & 0.0 & 349 & 0.342 & 0.760 & 0.337 & 0.697 \\
 & 0.0 & 1.0 & 0.5 & 349 & 0.375 & 0.777 & 0.364 & 0.720 \\
 & 0.4 & 0.4 & 0.2 & 349 & 0.401 & 0.809 & 0.390 & 0.761 \\
 & 0.5 & 0.3 & 0.15 & 349 & 0.386 & 0.809 & 0.371 & 0.762 \\
 & 0.6 & 0.2 & 0.2 & 349 & 0.379 & 0.810 & 0.359 & 0.762 \\
 & 0.6 & 0.4 & 0.0 & 349 & 0.342 & 0.792 & 0.327 & 0.742 \\
 & 0.6 & 0.4 & 0.2 & 349 & 0.390 & 0.810 & 0.375 & 0.762 \\
 & 0.6 & 0.4 & 0.4 & 349 & 0.403 & \textbf{0.811} & 0.388 & \textbf{0.763} \\
 & 0.6 & 0.4 & 0.6 & 349 & \textbf{0.406} & 0.809 & \textbf{0.390} & 0.762 \\
 & 0.6 & 0.5 & 0.3 & 349 & 0.400 & 0.810 & 0.387 & 0.762 \\
 & 0.6 & 0.6 & 0.2 & 349 & 0.394 & 0.808 & 0.383 & 0.760 \\
 & 0.7 & 0.5 & 0.25 & 349 & 0.392 & 0.810 & 0.377 & 0.762 \\
 & 0.8 & 0.4 & 0.2 & 349 & 0.381 & 0.808 & 0.364 & 0.760 \\
 & 1.0 & 0.0 & 0.0 & 349 & 0.236 & 0.715 & 0.207 & 0.661 \\
 & 1.0 & 0.4 & 0.2 & 349 & 0.368 & 0.806 & 0.350 & 0.758 \\
\midrule
\multirow{16}{*}{TS2Vec}
 & 0.0 & 0.0 & 1.0 & 349 & 0.355 & 0.713 & 0.328 & 0.662 \\
 & 0.0 & 1.0 & 0.0 & 349 & 0.342 & 0.760 & 0.337 & 0.697 \\
 & 0.0 & 1.0 & 0.5 & 349 & 0.375 & 0.777 & 0.364 & 0.720 \\
 & 0.4 & 0.4 & 0.2 & 349 & 0.470 & 0.866 & 0.448 & 0.827 \\
 & 0.5 & 0.3 & 0.15 & 349 & 0.473 & 0.870 & 0.447 & 0.830 \\
 & 0.6 & 0.2 & 0.2 & 349 & 0.475 & 0.871 & 0.447 & 0.830 \\
 & 0.6 & 0.4 & 0.0 & 349 & 0.468 & 0.868 & 0.441 & 0.826 \\
 & 0.6 & 0.4 & 0.2 & 349 & 0.473 & 0.869 & 0.447 & 0.829 \\
 & 0.6 & 0.4 & 0.4 & 349 & 0.474 & 0.867 & 0.448 & 0.827 \\
 & 0.6 & 0.4 & 0.6 & 349 & 0.473 & 0.863 & 0.447 & 0.822 \\
 & 0.6 & 0.5 & 0.3 & 349 & 0.472 & 0.867 & 0.448 & 0.827 \\
 & 0.6 & 0.6 & 0.2 & 349 & 0.472 & 0.868 & \textbf{0.449} & 0.828 \\
 & 0.7 & 0.5 & 0.25 & 349 & 0.473 & 0.869 & 0.448 & 0.829 \\
 & 0.8 & 0.4 & 0.2 & 349 & \textbf{0.475} & 0.871 & 0.448 & 0.830 \\
 & 1.0 & 0.0 & 0.0 & 349 & 0.470 & 0.869 & 0.436 & 0.823 \\
 & 1.0 & 0.4 & 0.2 & 349 & 0.475 & \textbf{0.871} & 0.447 & \textbf{0.831} \\
\midrule
\multirow{16}{*}{TSPulse$_{\mathrm{ZS}}$ encoder}
 & 0.0 & 0.0 & 1.0 & 349 & 0.355 & 0.713 & 0.328 & 0.662 \\
 & 0.0 & 1.0 & 0.0 & 349 & 0.342 & 0.760 & 0.337 & 0.697 \\
 & 0.0 & 1.0 & 0.5 & 349 & \textbf{0.375} & \textbf{0.777} & \textbf{0.364} & \textbf{0.720} \\
 & 0.4 & 0.4 & 0.2 & 349 & 0.256 & 0.722 & 0.218 & 0.658 \\
 & 0.5 & 0.3 & 0.15 & 349 & 0.232 & 0.715 & 0.194 & 0.650 \\
 & 0.6 & 0.2 & 0.2 & 349 & 0.225 & 0.709 & 0.186 & 0.645 \\
 & 0.6 & 0.4 & 0.0 & 349 & 0.179 & 0.692 & 0.143 & 0.624 \\
 & 0.6 & 0.4 & 0.2 & 349 & 0.237 & 0.716 & 0.199 & 0.652 \\
 & 0.6 & 0.4 & 0.4 & 349 & 0.256 & 0.718 & 0.218 & 0.655 \\
 & 0.6 & 0.4 & 0.6 & 349 & 0.268 & 0.719 & 0.231 & 0.656 \\
 & 0.6 & 0.5 & 0.3 & 349 & 0.253 & 0.720 & 0.215 & 0.656 \\
 & 0.6 & 0.6 & 0.2 & 349 & 0.246 & 0.721 & 0.209 & 0.656 \\
 & 0.7 & 0.5 & 0.25 & 349 & 0.240 & 0.717 & 0.202 & 0.653 \\
 & 0.8 & 0.4 & 0.2 & 349 & 0.224 & 0.712 & 0.185 & 0.647 \\
 & 1.0 & 0.0 & 0.0 & 349 & 0.107 & 0.586 & 0.068 & 0.519 \\
 & 1.0 & 0.4 & 0.2 & 349 & 0.213 & 0.709 & 0.174 & 0.644 \\
\bottomrule
\end{tabular}
\end{adjustbox}
\end{table}


\end{document}